\RequirePackage{etex} %% **************** add this for ARXIV submission
\documentclass[runningheads,twocolumn,a4paper,10pt]{llncs}
\usepackage{etex}

\reserveinserts{28}
\pdfoutput=1
\usepackage[a4paper, total={7in, 10in}]{geometry}

\usepackage{graphicx}
\usepackage{caption}
\usepackage{subcaption}
\usepackage{soul}
\usepackage[dvipsnames]{xcolor}
\usepackage{xcolor,listings}
\usepackage{textcomp}
\usepackage{todonotes}
\usepackage{array}
\usepackage{amsmath}
\usepackage{relsize}
\usepackage{amssymb}
\usepackage{color}
\usepackage{threeparttable}
\usepackage{hyperref}

\usepackage{algorithm}
\usepackage{algpseudocode}
\usepackage{tikz}
\usetikzlibrary{shapes.geometric, arrows,calc}
\tikzstyle{startstop} = [rectangle, rounded corners, minimum width=3cm, minimum height=1cm,text centered, draw=black, fill=none]

\tikzstyle{io} = [trapezium, trapezium left angle=70, trapezium right angle=110, minimum width=3cm, minimum height=1cm, text centered, draw=black, fill=blue!30]

\tikzstyle{process} = [rectangle, minimum width=3.5cm, minimum height=1cm, text centered, draw=black, fill=orange!0]

\tikzstyle{neuron} = [circle, minimum size=3cm, text centered, draw=black, fill=orange!0]

\tikzstyle{decision} = [diamond, minimum width=3cm, minimum height=1cm, text centered, draw=black, fill=green!30]

\tikzstyle{arrow} = [thick,->,>=stealth]

\usepackage{pgfplots}
\pgfplotsset{compat=newest}
\pgfplotsset{plot coordinates/math parser=false}
\usepackage{booktabs}
\usepackage{multirow}

\usepackage[maxnames=3,firstinits=true,sorting=none,doi=false,url=false,isbn=false]{biblatex}
\def \assumptionEWCfirst {1}
\def \assumptionEWCsecond {2}

\DeclareMathOperator*{\minimize}{min\text{ }}

\begin{document}

% \title{Operational Domain Adaptation using Regularisation Techniques}
\title{DIRA: \textbf{D}ynamic \textbf{I}ncremental \textbf{R}egularised \textbf{A}daptation}

% ====================
% Authors
% =====================
% ==== For Arxiv submission uncomment below
\titlerunning{Dynamic Domain Incremental ...}
\authorrunning{Ghobrial et al.}
\author{Abanoub Ghobrial$^{1,*}$, Xuan Zheng$^{1}$, Darryl Hond$^{2}$, Hamid Asgari$^{2}$, and Kerstin Eder$^{1}$}
\institute{University of Bristol, Bristol, UK$^{1}$ \\ Thales UK, Reading, UK$^{2}$}
\maketitle
\let\thefootnote\relax\footnotetext{
\noindent This research is part of an iCASE PhD funded by EPSRC and Thales UK. Abanoub Ghobrial (e-mail: abanoub.ghobrial@bristol.ac.uk), Xuan Zheng (e-mail: dq18619@bristol.ac.uk), and Kerstin Eder (e-mail: kerstin.eder@bristol.ac.uk) are with the Trustworthy Systems Lab, Department of Computer Science, University of Bristol, Merchant Ventures Building, Woodland Road, Bristol, BS8 1UB, United Kingdom. Darryl Hond (e-mail: darryl.hond@uk.thalesgroup.com) and Hamid Asgari (e-mail: hamid.asgari@uk.thalesgroup.com) are with Technology and Innovation Research, Thales, Reading, United Kingdom.
}

% ==== For IEEE submission 
% \author{Abanoub Ghobrial$^{1,*}$, Xuan Zheng$^{1}$, Darryl Hond$^{2}$, Hamid Asgari$^{2}$, and Kerstin Eder$^{1}$\\
% 	\normalsize $^{1}$ University of Bristol, Bristol, UK\\
% 	\normalsize $^{2}$ RTI, Thales UK, Reading, UK\\
% 	% \normalsize abanoub.ghobrial@bristol.ac.uk, dq18619@bristol.ac.uk, darryl.hond@uk.thalesgroup.com, hamid.asgari@uk.thalesgroup.com, kerstin.eder@bristol.ac.uk\\
% 	\normalsize *corresponding author

% \thanks{This research is part of an iCASE PhD funded by EPSRC and Thales UK. Abanoub Ghobrial (e-mail: abanoub.ghobrial@bristol.ac.uk), Xuan Zheng (e-mail: dq18619@bristol.ac.uk), and Kerstin Eder (e-mail: kerstin.eder@bristol.ac.uk) are with the Trustworthy Systems Lab, Department of Computer Science, University of Bristol, Merchant Ventures Building, Woodland Road, Bristol, BS8 1UB, United Kingdom. Darryl Hond (e-mail: darryl.hond@uk.thalesgroup.com) and Hamid Asgari (e-mail: hamid.asgari@uk.thalesgroup.com) are with Technology and Innovation Research, Thales, Reading, United Kingdom.
% }
% }
% \maketitle

% \author{\IEEEauthorblockN{Anonymous Authors}}

\thispagestyle{plain}
\pagestyle{plain}
\begin{abstract}

Autonomous systems (AS) often use Deep Neural Network (DNN) classifiers to allow them to operate in complex, high-dimensional, non-linear, and dynamically changing environments.
Due to the complexity of these environments, DNN classifiers may output misclassifications during operation when they face domains not identified during development.
Removing a system from operation for retraining becomes impractical as the number of such AS increases. 
To increase AS reliability and overcome this limitation, DNN classifiers need to have the ability to adapt during operation when faced with different operational domains using a few samples (e.g. 2 to 100 samples).
However, retraining DNNs on a few samples is known to cause catastrophic forgetting and poor generalisation. 
In this paper, we introduce Dynamic Incremental Regularised Adaptation (DIRA), an approach for dynamic operational domain adaption of DNNs using regularisation techniques. 
We show that DIRA improves on the problem of forgetting and achieves strong gains in performance when retraining using a few samples from the target domain. 
Our approach shows improvements on different image classification benchmarks aimed at evaluating robustness to distribution shifts (e.g.CIFAR-10C/100C, ImageNet-C), and produces state-of-the-art performance in comparison with other methods from the literature.  

\end{abstract}

% === For IEEE
% \begin{IEEEkeywords}
% Adaptation, Regularisation, Machine Learning
% \end{IEEEkeywords}
% \IEEEpeerreviewmaketitle 
% \IEEEpeerreviewmaketitle

\section{Introduction}
Autonomous systems (AS) often are developed using deep neural network (DNN) classifiers to interact and adapt in dynamically changing real-world environments to achieve their intended goals. 
The benefit of using DNNs in autonomous systems is their ability to learn complicated patterns 
% in high-dimensional data 
from complex environments, and thus produce highly non-linear decision boundaries to cope with the complexity of operational environments.
However, it is a challenge to verify the behaviour of DNNs.   
A popular example of such ASs is self-driving cars. Current research shows that for each self-driving car, an impractical amount of testing is required to verify the system for deployment~\cite{RR-1478-RC}. Innovative methods of increasing the efficiency of testing and validation are actively being developed to make the process more practical, e.g.~\cite{chance2020agency,eder2021complete}. 
However, due to the vast operational environments and the enormous effort required in testing to achieve deployment, the community is additionally incorporating trustworthiness assessment of AS to allow for reliable deployment and progressive improvements during the operational lifetime of such systems~\cite{chance2023assessing}. 
This follows a two-stage approach presented by Koopman~et.al.~\cite{Koopman2020}, which stipulates that, given an AS passes some minimum safety validation case, the system is deployed and is improved during operation to increase its reliability over time. 
Thus, the system is allowed to adapt to dynamically changing operational environments.

The process of continual learning or adaptation can be broken into three stages 1) detection of change in the operational domain, e.g.~\cite{Hond2020,Schaffer2017, Mandelbaum2017, Xing2019}, 2) supply of labels from an oracle or ground truth for new operational domain samples, e.g.~\cite{Barr2015,Zhang2020}, 3) retraining. 
Alternatively, 2) and 3) can be substituted by one stage of self-supervised or unsupervised retraining. We aim to investigate this option in future work.
In this paper, we focus on point 3), i.e. retraining.

DNN classifiers use gradient-based optimisation algorithms to learn. 
The gradient optimiser modifies the decision boundary based on the samples used in training. 
Retraining using few samples can result in a phenomenon known as \textit{catastrophic forgetting}, where the model overfits to the few training samples used and does not generalise to the domain distribution~\cite{Goodfellow2014}. 
Generally, to overcome catastrophic forgetting, new samples are added to the initial training dataset and the classifier is fully retrained. 
Full retraining, however, can be cumbersome to perform during operation. 
In this paper, we propose Dynamic Incremental Regularised Adaptation (DIRA), a framework to achieve operational domain adaption by retraining using only few samples from the target domain. 
We utilise a combination of regularisation techniques and a retraining scoring approach in our framework to overcome the need for full retraining.

Practically, upon adaptation of AS, reassessment of the system's safety may be required.  
The safety compliance of evolving DNN classifiers during operation against a set of requirements or regulations is beyond the scope of this paper, but may be achieved through the use of runtime safety behavioural checkers as presented by Harper~et.al.~\cite{harper2021safety} or by using online methods for quantifying trustworthiness in predictions during operation as shown by Ghobrial et al~\cite{ghobrial2023trustworthiness}. 

In the next section, we discuss related work material. Section~\ref{sec:method} introduces our method.
Experimentation Setup and Results \& Discussion are handled by sections~\ref{sec:experimentation} and \ref{sec:results}, respectively.
We conclude and discuss future works in Section~\ref{sec:conclusions}.

\section{Related work}
\subsection{Types of Incremental Learning}
Gradually assimilating new information from a continuously changing data stream, known as `continual learning', poses a challenge for deep neural networks. 
Continual learning, however, is a fundamental aspect of evolving autonomous systems.
In a continual learning setting the problem is broken down into several parts that need to be learned sequentially.
In the continual learning literature, these parts are often called \textit{tasks}. Thus, the term tends to have several meanings. 
These several connotations of the term task make it difficult to study the different challenges associated with continual learning. 
To overcome this problem, Ven et al.\cite{Ven2022}, proposed to brake down continual learning into three incremental learning scenarios: task-incremental, domain-incremental, and class-incremental learning (see Table~\ref{tab:incremental_types}). 
Each scenario describes the context of the parts required to be learned sequentially, formerly the three scenarios contexts were referred to using the term task.
Braking continual learning into different scenarios makes it more convenient to study the different challenges associated with each scenario, and subsequently develop appropriate techniques to overcome the associated challenges~\cite{li2022energy,lesort2021understanding,zeno2018task,gepperth2016incremental}. 

The first scenario (task-incremental learning), describes the case where the algorithm is required to learn incrementally a set of distinct tasks.
For example, if a neural network model was to classify numbers from 0 - 9 in English (like in the MNIST dataset~\cite{deng2012mnist}), then a new task for the model can be to learn to classify samples in Permuted-MNIST~\cite{Goodfellow2014} or Fashion-MNIST~\cite{Xiao2017} i.e. the same number of classes but the pattern has changed distinctively. For more examples see~\cite{ramesh2021model, masse2018alleviating, ruvolo2013ella}.
In the second scenario (domain-incremental learning), the model needs to learn the same problem but in different contexts, because the domain or input distribution has shifted. Using our previous example of classifying digits 0 - 9, in domain incremental learning, the model is required to learn to classify digits 0 - 9 but with Gaussian noise or contrast noise added to the input samples. See~\cite{JehanzebMirza2022, ke2021classic} for more examples.
The third scenario (class-incremental learning), describes when the model needs to learn a growing number of classes. In the example of classifying digits 0 - 9, class-incremental learning is the model learning to classify digit `10' as an additional class to the existing 0 - 9 classes. See examples \cite{Tao2020, rebuffi2017icarl}.

Since our focus is on dynamic distributional shifts during operation, we are interested in the second scenario of incremental learning i.e. domain-incremental learning. We focus on trying to achieve domain incremental adaptation using a limited number of samples.

\begin{table}[]
    \centering
    \begin{tabular}{l|l}
         \textbf{Scenario} & \textbf{Description}  \\ \hline
         Task-Incremental & Sequentially learn to solve a number \\
         Learning & of distinct tasks.\\\hline
         
         Domain-Incremental & Learn to solve the same problem in \\
         Learning & different contexts.\\ \hline
         Class-Incremental & Differentiate between incrementally\\
         Learning & observed classes. \\\hline
         
    \end{tabular}
    \caption{Overview of incremental learning scenarios~\cite{Ven2022}}
    \label{tab:incremental_types}
\end{table}

\subsection{Domain Adaptation Frameworks}
There have been a number of introduced approaches in the literature that address the problem of domain-incremental adaptation.
%, specifically in dynamic setups where only a limited number of samples from the new domain are available for retraining.
%
For a breakdown of categories for the different introduced approaches in the literature, we direct interested readers towards~\cite{mirza2022norm}.
Here we will cover some state-of-the-art examples of these approaches relevant to our results discussed later in section~\ref{sec:results}.
% TTT~\cite{sun2020test}, NORM \cite{schneider2020improving, nado2020evaluating}, DUA~\cite{mirza2022norm}.

One popular approach is using self-supervision to achieve domain adaptation. Test-time training (TTT) combine different self-supervised auxiliary contexts to achieve domain adaptation. They break down neural network parameters into three parts, such that pictorially the architecture has a \textit{Y}-structure. 
The bottom section of the Y-structured architecture represents the input layer and the layers responsible for the shared feature extraction, whilst the other two sections contain layers for learning and outputting labels for the main and auxiliary tasks independently. An example of this auxiliary task is being able to tell the rotation of the input image. 
During training time the whole neural network is optimised using a combined loss function that aims to maximise performance on both the main and auxiliary tasks. 
During retraining to adapt to a new domain, only parameters of the shared feature extraction and the auxiliary task sections are allowed to change. 
By doing so the the shared feature extraction section of the network modifies to learn the new domain, so then the network may output correct predictions on the unchanged branch of the network responsible for the main task~\cite{sun2020test, sun2019unsupervised}.

Correcting domain statistics is another common approach to achieving domain adaption, e.g.~\cite{mirza2022norm, schneider2020improving, nado2020evaluating, maria2017autodial}. 
%
% Specifically, results for correcting statistics of batch normalisation layers~\cite{ioffe2015batch} have shown competitive results for self-supervision approaches such as TTT.
%
Some of these approaches rely on using a large number of samples to recalculate the running mean and standard deviation of batch normalisation layers for the target domain e.g.~\cite{nado2020evaluating,schneider2020improving}.
Other approaches, like Dynamic Unsupervised Adaption (DUA)~\cite{mirza2022norm}, combine the running mean and standard deviation for normalisation layers from the original domain and the target domain to achieve adaptation in an unsupervised fashion, whilst using significantly fewer samples (typically $\approx100$ samples). 

Our proposed DIRA method aims at achieving adaptation through the regularisation of new and old information.%approach also relies on correcting the domain statistics but varies in several ways from the described approaches.
We retrain the model on samples from the target domain. %, similar to how the model was initially trained, i.e. by tweaking the model parameters using optimisation techniques, such as stochastic gradient descent. 
Therefore, we require labels to be provided with the retraining samples, which makes our approach a supervised instead of an unsupervised method. 
We use regularisation techniques to avoid catastrophic forgetting and achieve adaptation using very few samples. 
By doing so, we benefit from transfer learning of information from the initial domain to the target domain.
Our philosophy is that if humans use transfer learning to learn and adapt to different environments, we believe that neural networks can also achieve domain adaptation in a similar fashion.
We see our approach can be combined with self-supervision methods, such as done in TTT, to overcome the need for providing labels. 
Exploring the use of self-supervision in our approach is left as future work. In this paper, we assume labels for samples from the target domain are available for retraining. 

\subsection{Regularisation}
The concept of regularisation allows a neural network to learn new information whilst retaining previously learned information. 
This allows a neural network to learn new information without experiencing catastrophic forgetting and without needing access to training data of previously learnt information.
Regularisation achieves this by presenting a penalisation term in the loss function of the optimisation problem. 
Several works in the literature have introduced different penalisation terms, some popular examples are Synaptic Intelligence (SI)~\cite{Zenke2017}, Learning without forgetting (LWF)~\cite{Li2018c}, and Elastic Weight Consolidation (EWC)~\cite{kirkpatrick2017overcoming}.  
In DIRA, different regularisation techniques may be utilisable, however, based on surveys such as the one provided by Kemker~\cite{Kemker2018a}, the EWC penalisation term results in state-of-the-art performance within regularisation techniques. Therefore, we developed our method predominantly based on EWC.

\section{Method} \label{sec:method}
We first summarise EWC regularisation~\cite{kirkpatrick2017overcoming} as our approach revolves around it. Then we discuss the details of our DIRA method.

\subsection{Elastic Weight Consolidation}\label{sec:EWC}
Kirkpatrick et al.~\cite{kirkpatrick2017overcoming} introduced Elastic Weight Consolidation (EWC) to overcome forgetting in task-incremental learning. 
EWC overcomes forgetting by introducing a penalisation term in the loss function when retraining. 
This penalisation term provides a sense of the importance of each weight in the trained model on the original classification task. 
Therefore, when retraining on a new task, the algorithm is guided to avoid making significant changes to weights with high importance to the initial task.
In this paper, we are interested in adapting to new domains, instead of adapting to new tasks. 
In the rest of this section, we will discuss the derivation of EWC and outline the assumptions that need to be taken into account when using EWC for domain adaptation. 

% ========== Derivation

During training of a DNN, the goal is to minimise the loss function $\mathcal{L}(\theta)$, represented as the Log-Likelihood function $-log(P(\theta|D))$~\cite{GoodBengCour2016}.
This aims at estimating $\theta$, which is the set of weights and biases in a DNN, given $D$, the dataset representing the samples of the distribution of interest.
$D$ can be split into two independent datasets such that $D=\{D_A, D_B\}$, where $D_A$ and $D_B$ are datasets that are trained on sequentially and each of them may represent a different distribution. 
Using the chain rule in probability it can be shown that: 

\begin{equation}\label{eq_loss_fucntion}
\begin{split}
log(P(\theta|D_A,D_B))&=log(P(D_B|\theta,D_A))+log(P(\theta|D_A)) 
\\
& - log(P(D_B|D_A))
\end{split}
\end{equation}
Considering the RHS of equation~\ref{eq_loss_fucntion}:
\begin{itemize}
    \item First term, using conditional independence $log(P(D_B|\theta,D_A)) = log(P(D_B|\theta))$ and hence can be seen as the loss function, $\mathcal{L_B}(\theta)$ that needs to be minimised for the new distribution or dataset $D_B$ alone.
    \item Second term, $log(P(\theta|D_A))$ is the loss function for training the neural network on distribution $D_A$ only. Thus can be denoted as $\mathcal{L_A}(\theta)$. 
    \item The Third term, is irrelevant as this term is constant with respect to $\theta$ and thus is lost when optimising using the stochastic gradient descent (SGD) i.e. does not need to be computed. We will neglect this term for the rest of the derivation.
\end{itemize}
Therefore, the overall loss function in equation \ref{eq_loss_fucntion} can be written as:
\begin{equation}\label{eq_loss_fucntion_short}
\mathcal{L}(\theta)=\mathcal{L_B}(\theta)+\mathcal{L_A}(\theta)
\end{equation}

In continual learning, distribution $A$ would have been trained on initially, and later samples from distribution $B$ arise and must be learnt by the DNN.
In this case, the term $\mathcal{L_A}(\theta)$ is considered to be intractable as it is assumed that access to training samples for distribution A is not available after initial training.

The underlying idea of EWC is to take a Bayesian approach to adapt the DNN model parameters, therefore learning additional distributions whilst avoiding catastrophic forgetting or minimising forgetting. 
However, due to intractable terms, it is not possible to maintain the full posterior $P(\theta|D))$.
An inference technique is required to approximate these intractable terms. 
EWC can be seen as an online approximate inference algorithm~\cite{Huszar2018}.
An essential assumption for EWC to approximate $\mathcal{L_A}(\theta)$ is that the DNN has been optimised for $D_A$ such that $\theta$ has reached a local or a global minimum, $\theta^{*}_{A}$, for distribution $D_A$.
This allows for $-log(P(\theta|D_A))$ to be approximated as a Gaussian distribution function at its mode using Laplace's method~\cite{MacKay2003}. 
Expanding $-log(P(\theta|D_A))$ using Taylor series around $\theta^{*}_{A}$: 

% log(P(\theta|D_A,D_B))&=log(P(D_B|\theta,D_A))+log(P(\theta|D_A)) 
% \\
% & - log(P(D_B|D_A))
\begin{equation} \label{L_A_exapnsion}
\begin{split}
    -log(P(\theta|D_A)) \approx & -log(P(\theta^*_A|D_A)) 
    \\
    & + (\frac{\partial(-log(P(\theta|D_A)}{\partial \theta}\vert_{\theta^*_A})(\theta - \theta^*_A) 
    \\
    & + \frac{1}{2}(\theta - \theta^*_A)^{T} H(\theta^*_A)(\theta - \theta^*_A) 
    \\
    & + \cdots
\end{split}
\end{equation}

\noindent Considering the RHS of equation~\ref{L_A_exapnsion}:
\begin{itemize}
    \item The First term, is a constant and similar to earlier it will get lost in the SGD optimiser.  
    \item Second term, evaluates to gradient 0 as it is assumed that $\theta^*_A$ is at the mode of the distribution. 
    \item  Third term; $H(\theta^{*}_{A})$ is the Hessian of $-log(P(\theta|D_A))$ with respect to $\theta$ evaluated at $\theta^*_A$, which is $(\frac{\partial^2(-log(P(\theta|D_A)}{\partial \theta^2}\vert_{\theta^*_A})$.  
\end{itemize}

The Hessian can be computed by approximating it to the empirical Fisher information matrix.
Using Bayesian rule: 
\begin{equation} \label{Bayesian_Approximation_To_Gaussian}
\begin{split}
    H(\theta^*_A) = &-\frac{\partial^2(log(P(D_A|\theta)))}{\partial\theta^2}\Bigg|_{\theta^*_A}
    \\
    & - \frac{\partial^2(log(P(\theta)))}{\partial\theta^2}\Bigg|_{\theta^*_A} 
    \\
    & + \frac{\partial^2(log(P(D_A)))}{\partial\theta^2}\Bigg|_{\theta^*_A}
\end{split}
\end{equation}

\noindent Considering the RHS of equation~\ref{Bayesian_Approximation_To_Gaussian}:
\begin{itemize}
    \item First term can be approximated as the negative of the empirical Fisher information matrix, $F$,~\cite{Kay1993,Ly2017,Martens2020}. 
    The Fisher matrix can be defined as a way of measuring the amount of information that a random observation $D_A[n]$ carries about a set of unknown parameters $\theta$ of a distribution that models $log(P(D_A|\theta)$, where $n$ is an index falling within the size, $N$, of the observable random samples $D_A$. 
    Formally, it is the negative of the expected value of the observed information, hence it can be shown that it approximates to the first term of equation~\ref{Bayesian_Approximation_To_Gaussian}: 
    \begin{equation} \label{approximation_to_fisher}
    \begin{split}
    F(\theta) &= -N E\left[\frac{\partial^2(log(P(D_A[n]|\theta)))}{\partial\theta^2}\right]
    \\
    & \approx -N \frac{1}{N}\mathlarger{\mathlarger{\sum}}_{n=1}^{N} \frac{\partial^2(log(P(D_A[n]|\theta)))}{\partial\theta^2}
    \\
    &
    = -\mathlarger{\mathlarger{\sum}}_{n=1}^{N} \frac{\partial^2(log(P(D_A[n]|\theta)))}{\partial\theta^2} 
    \\
    &
    = -\frac{\partial^2(log(P(D_A|\theta)))}{\partial\theta^2} 
    \end{split}
    \end{equation}
    
    \noindent The approximation made to the expectation in equation~\ref{approximation_to_fisher} becomes exact as the number of observations or samples becomes infinite.
    Therefore, the data size $N$ of the previous information is crucial to the applicability of using the EWC approximation.

    \item Second term, is the \textit{prior probability}.That is the probability distribution the DNN represents before being trained on any observations i.e. datasets.
    Given that often $\theta$ in DNNs are initialised using a random uniform distribution, then this term evaluates to zero and hence is ignored by the EWC algorithm.

    \item Third term, evaluates to zero as non-dependent on $\theta$.
    
\end{itemize}

\noindent Putting terms together from the previous steps makes equation~\ref{eq_loss_fucntion_short} reach the EWC loss function presented by ~\cite{kirkpatrick2017overcoming}: 
\begin{equation}\label{EWC_fucntion}
   \mathcal{L}(\theta) = \mathcal{L}_{B}(\theta) +  \sum_{j}^{} \frac{\lambda}{2} F_{A,j} (\theta_j - \theta^{*}_{A,j})^2
\end{equation}
where $\lambda$ is a hyper-parameter presented by Kirkpatrick et al. to allow for fine-tuning to minimise forgetting, and $j$ labels each parameter.
\noindent We summarise the list of assumptions for which equation~\ref{EWC_fucntion} holds:

\noindent\textbf{Assumption \assumptionEWCfirst:} The DNN was trained very well on the previous distribution represented by $D_A$ that $\theta$ has reached a local or a global minimum i.e. $\theta^{*}_{A} = argmin_{\theta}\{ -log(P(\theta|D_A))\}$. 

\noindent\textbf{Assumption \assumptionEWCsecond:} ``Enough'' observations are available in $D_A$ to allow for the approximation from the Hessian to the empirical Fisher information matrix.

% ===================================================
\subsection{Dynamic Incremental Regularised Adaptation (DIRA)}
This section describes the algorithmic details of our method.
In order to achieve successful domain-adaptation we have taken into consideration the two assumptions outlined in section~\ref{sec:EWC} when developing DIRA.
Let $M_0$ be the model trained on the original domain dataset $X_0$. 
The standard optimisation problem in training a neural network on the original domain with a loss function $\mathcal{L}_0$ solves:
\begin{equation}
    \displaystyle{\minimize_{\theta}} \mathcal{L}_0(\theta)
\end{equation}

The aim of our approach is to adapt the trained model to out-of-distribution target data $X_{T}$ using a few number of samples $S_T$ from the target domain.
To achieve this goal we utilise the concept of transfer learning, aiming at reserving beneficial information learnt from the original domain to allow for successful adaptation to the target domain. 
Our hypothesis is that by using regularisation techniques one should be able to utilise this notion of transfer learning to achieve adaption with a limited number of samples from the target domain. 
Therefore, the problem we try to optimise for during adaptation becomes a combination of the loss function for the original domain $\mathcal{L}_0$ and the target domain $\mathcal{L}_T$: 
\begin{equation}
    \displaystyle{\minimize_{\theta}} \mathcal{L}_T(\theta) 
+ \mathcal{L}_0(\theta)
\end{equation}

The $\mathcal{L}_0$ is intractable during adaptation since we have no access to the original domain training data. 
Therefore, an approximation of the original domain is done using EWC which yields the optimisation problem:

\begin{equation}
    \displaystyle{\minimize_{\theta}} \mathcal{L}_T(\theta) 
+ \sum_{j}^{}\lambda F_{0,j}(\theta_j - \theta^{*}_{0,j})^2
\end{equation}

To satisfy assumption 1, whenever we retrain we always start from the original model $M_0$. 
Practically this is achievable as a copy of $M_0$ can always be kept onboard of a system. Assumption 2 can be satisfied by calculating the Fisher matrix using the original training dataset during initial training and a copy of this calculated Fisher matrix would be saved on board of the system, omitting the need to keep a copy of the initial training data on board.

In each training step $t$, the model parameters are updated according to Equation~\ref{theta_new_substituted}, where $\eta$ is the learning rate. 
\begin{equation}
   \theta_{t+1} = \theta_{t} - \eta \bigg( \dfrac{\mathcal{L}_{\text{T}}(\theta)}{d\theta} - 
   \dfrac{\sum_{j}^{}\lambda F_0(\theta_{t,j} - \theta^{*}_{0,j})^2}{d\theta} \bigg)
   \label{theta_new_substituted}
\end{equation}

% - Dicuss the interplay between learning rate and lambda
The two hyperparameters critical for the success of our optimisation problem are $\eta$ and $\lambda$. 
Different numerical search methods can be used for finding values for these hyperparameters, e.g. grid search, Bayesian optimisation etc.
From empirical testing, we found that a combination of $\eta = 1$e-5 and $\lambda=1$ yields near optimum adaptation for datasets we used in our experimentation.
% However, an interplay between the values of these hyperparameters is needed as the number of retraining samples from the target domain increases. This is to avoid unstable retraining or inefficient adaptation.
%
% Therefore a method for online evaluation of the retrained model during operation is needed.
%
% We propose to use a scoring approach that ranks the efficiency of the adaptation based on the model's accuracy on the original and target domains.
%
% During operation the original domain training data is inaccessible, however, we assume some test data, $X_{0, test}$ for the original domain is accessible to the system during operation. 
%
% For the target domain, we only have access to samples, $S_{T}$ used in retraining, which can be used as our test data for the target domain when calculating a score that ranks the efficiency of adaptation.  
%
% Now to rank the efficiency of adaptation, we propose a simple scoring approach, Controlled Adaptation Score (CAS), shown by Equation~\ref{eq:CAS_score}. 
%
% $A_0$ is the retrained model accuracy on the test dataset $X_{0,test}$ and $A_T$ is the retrained model accuracy on the samples $S_{T}$. 
%
% $\zeta$ is a constant. Through empirical testing, we found that a value of $\zeta=10$ yields near-optimum adaptation using DIRA.
%
% The higher the CAS, the better the adaption efficiency. Using CAS, appropriate values for $\lambda$ and $\eta$ can be found during operation as the number of available samples from the target domain increases.

% \begin{equation}
% CAS = A_T + \zeta \cdot A_0  
% \label{eq:CAS_score}
% \end{equation}

\section{Experimentation Setup}\label{sec:experimentation}
We used the problem of image classification to showcase our method. All of our experimentation was based in PyTorch library~\cite{paszke2019pytorch}. In the rest of this section, we discuss the details of our experimentation setup. Code is available at this repository: \url{https://github.com/Abanoub-G/DIRA}

\subsection{Benchmarks}
We utilise CIFAR-10C, CIFAR-100C, and ImageNet-C datasets in our experimentation. These are image classification benchmarks to evaluate a model's robustness against common corruptions~\cite{hendrycks2019benchmarking}.
The benchmarks add different corruptions to the tests sets of CIFAR-10/CIFAR-100\cite{krizhevsky2009learning} and ImageNet~\cite{deng2009imagenet}
There are 20 corruptions in total with five different levels of severity, however, most SOTA domain-incremental retraining frameworks utilise 15 corruptions out of the 20 in their comparisons, e.g.~\cite{mirza2022norm, sun2020test}. 
These are deemed the more common corruptions. 
We use the same 15 common corruptions used by other methods in the literature.
% When we compare with other frameworks we use the same 15 common corruptions, otherwise, we use all 20 corruptions provided.
% \todo[inline]{Add CIAFRR100C and Imagenet-C after you finish experimetns }

\subsection{Baselines}
We list below the different baselines we assess against our DIRA approach:
\begin{enumerate}
    \item \textbf{Source}: Refers to results of the corresponding baseline model trained on the incorrupt data (i.e. $X_0$), without adaption to the target domain.

    \item \textbf{SGD}: Denotes retraining on samples of the corrupt data using only Stochastic Gradient Decent optimisation~\cite{GoodBengCour2016}, i.e. without using any complimentary incremental learning frameworks, similar to how initial training on the incorrupt data is done.

    \item \textbf{TTT}~\cite{sun2020test}: Test-Time Training (TTT) adapts parameters in the initial layers of the network by using auxiliary tasks to achieve self-supervised domain adaption. 
    
    \item \textbf{NORM}~\cite{schneider2020improving, nado2020evaluating}: Ignores the initial training statistics and recalculates the batch normalization statistics using samples from the target domain only (requiring a large number of samples).

    \item \textbf{DUA}~\cite{mirza2022norm}: Dynamic Unsupervised Adaptation (DUA), takes into account initial training statistics and updates batch normalization statistics using samples from the target domain (requiring few samples).
    
\end{enumerate}

\subsection{Models and Hardware}
We used ResNets~\cite{he2016deep} in our experiments, utilising two versions of ResNet: ResNet-18 (18-layer) and ResNet-26 (26-layer). For CIFAR-10/CIFAR-100, we used ResNet-26. Initial training for the models was done locally. For ImageNet, we used a pre-trained off-the-shelf ResNet-18 model from PyTorch~\cite{paszke2019pytorch}. 
Experiments for CIFAR10 and CIFAR100 were done on an MSI GF65 THIN 3060 Laptop with 64GB RAM and a Linux Ubuntu 20.04.2 LTS (64-bit) operating system, whilst for ImageNet we used a Dell Alienware Desktop PC with 64GB RAM and a Linux Ubuntu 18.04.4 LTS (64-bit) operating system.

To achieve reliable comparisons against baselines the starting model parameters from which retraining is done must be the same.
Otherwise, the accuracy improvements cannot be reliably attributed to the effectiveness of the retraining method and can be argued that it is due to varying starting model accuracies or parameters.
Therefore, in our results for CIFAR10/100 on ResNet-26 we only compare against Source, as we do not have the initial models used in retraining by other SOTA methods.
For ImageNet on ResNet-18 we compare against SOTA methods because the starting trained model used by other SOTA retraining methods is the same off-the-shelf ResNet-18 model from PyTorch.

\subsection{Optimisation Details}
We used Stochastic Gradient Decent (SGD) for optimisation during training and retraining in our work.
For retraining DIRA, we found from empirical testing that $\eta = 1$e-5 and $\lambda=1$ yields near optimum adaptation. %to select values for $\lambda$ and $\eta$ from a specified sets of values of $[0.25, 0.5, 0.75, 1, 2]$ and $[1\text{e-}5,1\text{e-}4,1\text{e-}3,1\text{e-}2]$ respectively. 
%
% We used a naive approach of grid search to find the optimal values to use when retraining. 
%
% These optimal values were based on which combination yields the highest CAS score.
%
The retraining is relatively quick as only a small number of samples are used and the retraining is done over 10 epochs.
%
% If one is interested in increasing the retraining speed, one may want to try other tuning approaches, like Bayesian optimisation~\cite{Mockus1975,Snoek2012,Feurer2019}, to find suitable values for $\lambda$ and $\eta$ faster.
%
% Through empirical testing, we found that a value for $\zeta=10$ from equation~\ref{eq:CAS_score} yields optimal performance for DIRA in our experimentation. 
%
We use top-1 classification accuracy as our assessment metric in all experiments~\cite{ghobrial2023evaluation}.

% \begin{figure}[]
%     \centering
%     \includegraphics[width=0.51\textwidth]{other/figures/CIFAR10C_Resnet18_N_TvsA_T_v2.pdf}
%     \caption{ResNet-18 mean classification accuracy over 15 different corruption types on CIFAR-10C at the highest severity (Level 5).}
%     \label{fig:resnet18_samplesVSaccuracy}
% \end{figure}

\begin{figure}[]
    \centering
    \includegraphics[width=0.51\textwidth]{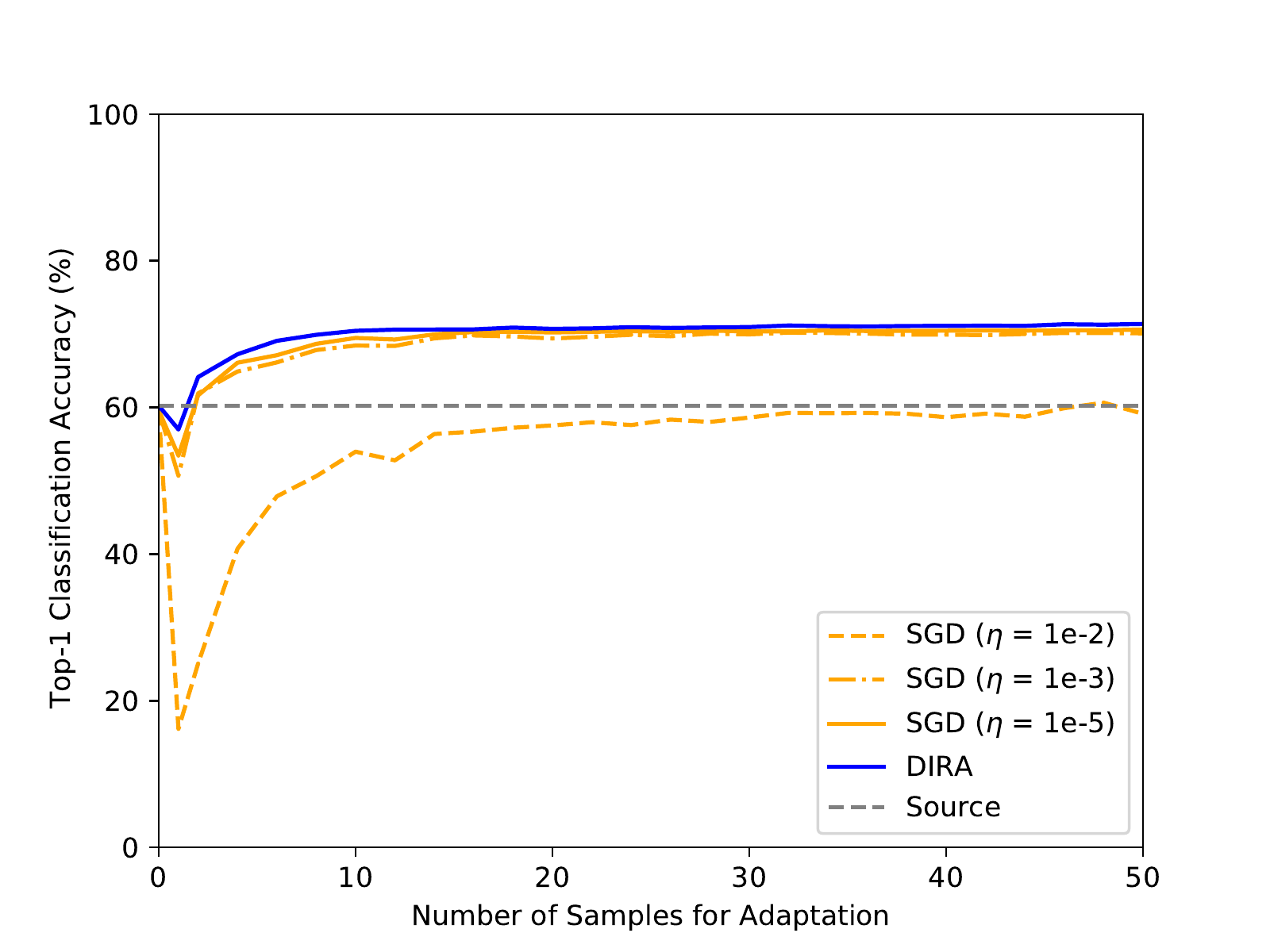}
    \caption{ResNet-26 mean classification accuracy over 15 different corruption types on CIFAR-10C at the highest severity (Level 5).}
    \label{fig:resnet26_samplesVSaccuracy}
\end{figure}

\begin{figure}[]
    \centering
    \includegraphics[width=0.51\textwidth]{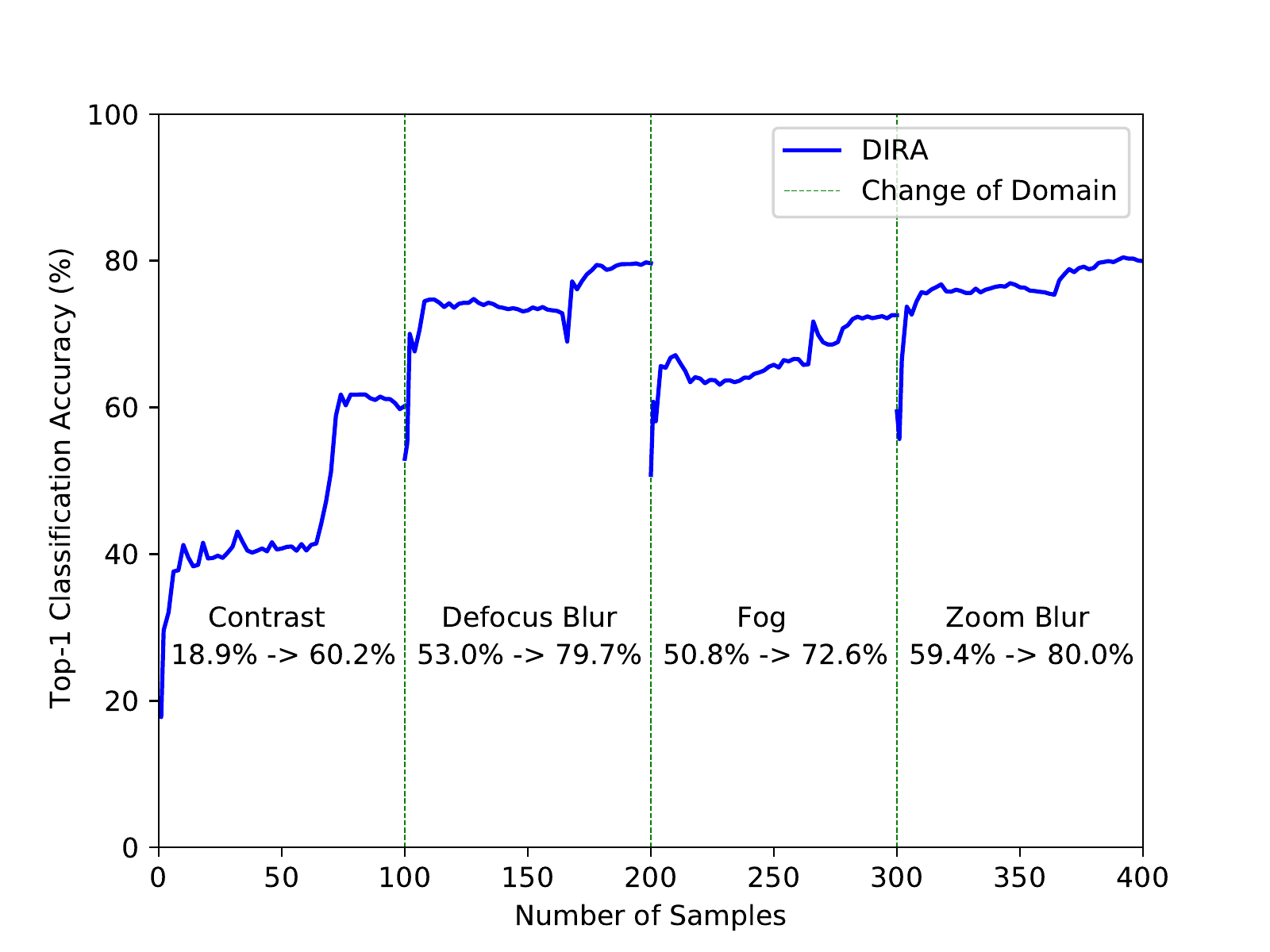}
    \caption{Dynamic adaptation scenario example for DIRA to different domains from CIFAR-10C. Pre-trained ResNet-26 on CIFAR-10 adapts to different corruption examples from CIFAR-10C dataset at the highest severity (Level 5), to show how well DIRA can dynamically adapt to operational domains.}
    \label{fig:DynamicScenario}
\end{figure}

\begin{table*}[]
    \centering
    \resizebox{\textwidth}{!}{\begin{tabular}{c|c c c c c c c c c c c c c c c|c}
                 & gaus & shot & impul & defcs & gls & mtn & zm & snw & frst & fg & brt & cnt & els & px &jpg & mean \\\hline
Source & 58.5 & 61.3 & 37.3 & 51.9 & 59.6 & 58.6 & 58.1 & 73.3 & 67.8 & 50.0 & 80.7 & 19.2 & 71.8 & 66.1 & 79.8 & 59.6 \\
% DIRA  & 75.0 & 75.7 & 52.5 & 79.6 & 66.4 & 76.1 & 79.2 & 77.7 & 77.4 & 71.0 & 84.4 & 61.6 & 74.0 & 77.9 & 79.4 & 73.9 \\ 
DIRA  & 73.6 & 75.6 & 61.9 & 79.7 & 65.8 & 77.9 & 80.0 & 77.4 & 77.0 & 72.6 & 84.2 & 60.2 & 74.9 & 76.9 & 79.5 & 74.5 \\
    \end{tabular}}
    \caption{Top-1 Classification Accuracy (\%) for each corruption in CIFAR-10C at the highest severity (Level 5). Source shows the results from the same model trained on the clean train set (CIFAR-10) and tested on the corrupted test set (CIFAR-10C). ResNet-26 is used with 100 retraining samples.}
    \label{tab:DIRA_CIFAR-10C_results}
\end{table*}

\begin{table*}[]
    \centering
    \resizebox{\textwidth}{!}{\begin{tabular}{c|c c c c c c c c c c c c c c c|c}
                 & gaus & shot & impul & defcs & gls & mtn & zm & snw & frst & fg & brt & cnt & els & px &jpg & mean \\\hline
Source  & 24.2 & 27.0 & 9.7 & 30.0 & 30.9 & 33.6 & 35.5 & 38.8 & 34.6 & 19.6 & 44.8 & 8.4 & 43.4 & 39.8 & 50.0 & 31.4 \\
DIRA  & 44.7 & 45.1 & 33.6 & 50.9 & 40.4 & 49.6 & 52.3 & 47.3 & 46.6 & 37.9 & 55.2 & 33.3 & 47.0 & 51.5 & 51.7 & 45.8 \\ 

    \end{tabular}}
    \caption{Top-1 Classification Accuracy (\%) for each corruption in CIFAR-100C at the highest severity (Level 5). Source shows the results from the same model trained on the clean train set (CIFAR-100) and tested on the corrupted test set (CIFAR-100C). ResNet-26 is used with 100 retraining samples.}
    \label{tab:DIRA_CIFAR-100C_results}
\end{table*}

\begin{table*}[]
    \centering
    \resizebox{\textwidth}{!}{\begin{tabular}{c|c c c c c c c c c c c c c c c|c}
                 & gaus & shot & impul & defcs & gls & mtn & zm & snw & frst & fg & brt & cnt & els & px &jpg & mean \\\hline
Source  & 1.6 & 2.3 & 1.6 & 9.4 & 6.6 & 10.2 & 18.2 & 10.5 & 15.0 & 13.7 & 48.9 & 2.8 & 14.7 & 23.1 & 28.3 & 13.8\\
TTT  & 3.1 & 4.5 & 3.5 & 10.1 & 6.8 & 13.5 & 18.5 & 17.1 & 17.9 & 20.0 & 47.0 & \textbf{14.4} & 20.9 & 22.8 & 25.3 & 16.4\\
NORM  & \textbf{12.9} & 10.4 & 9.5 & \textbf{12.4} & 10.6 & \textbf{20.0} & 28.1 & \textbf{29.4} & 18.5 & 33.1 & 52.2 & 10.2 & 26.5 & 35.8 & 31.5 & 22.7\\
DUA  & 10.6 & 12.4 & \textbf{11.9} & 12.0 & 11.4 & 15.3 & 25.7 & 22.2 & 21.6 & 31.4 & 54.4 & 4.1 & 27.8 & 33.5 & \textbf{32.6} & 21.8\\
DIRA  & 12.0 & \textbf{13.5} & 11.6 & 10.2 & \textbf{11.5} & 18.7 & \textbf{31.2} & 26.6 & \textbf{27.2} & \textbf{36.3} & \textbf{56.3} & 9.2 & \textbf{35.7} & \textbf{38.1} & 32.0 & \textbf{24.7}\\ 
    \end{tabular}}
    \caption{Top-1 Classification Accuracy (\%) for each corruption in ImageNet-C at the highest severity (Level 5). Source shows the results from the same model trained on the clean train set (ImageNet) and tested on the corrupted test set (ImageNet-C). For a fair comparison with TTT, NORM, and DUA, we use the same initially trained ResNet-18 model. 100 retraining samples are used. Highest accuracy is highlighted in bold.}
    \label{tab:DIRA_SOTA_ImageNet_results}
\end{table*}

\section{Results \& Discussion}
\label{sec:results}
% \todo[inline]{Add discussion point saying that we provide results for CIFAR10 and CIRA100 as suplimentary material. The results for CIAR10 and CIRA100 is to show the effectiveness of DIRA on other datasets, however it is not to be relied upon when comparing with SOTA techniques, as the starting points for the model may be different and the model architecture used may be different. For instance, inseccpting the code provided by the DUA method for Resnet26, we find that three blocks of the network were used instead of four layers. Whilst in the official implementation provided by Pytorch there are four layers. We base our experiments on the official implementation. In comparing with SOTA methods we ustilise REsnet-18 and Iamgnet1K, as this has a standardised pretrained model provided by PyTorch. Methods we compare with, utilise this stadnardised model in their case studies. Therefore we focus on these set of results as our main comparison.}

% In this section we present our detailed results, showing how DIRA tackles catastrophic forgetting, showcase an adaptation scenario for DIRA dealing with different domains consecutively, and lastly comparing our DIRA approach with other SOTA adaptation methods from the literature.

\subsection{Regularisation effect on adaptation}
To investigate overall adaptation improvement using regularisation we plot Figure~\ref{fig:resnet26_samplesVSaccuracy}. 
Figure~\ref{fig:resnet26_samplesVSaccuracy} shows how top-1 classification accuracy changes as the number of samples available from the target domain increases. 
The naive approach would be to retrain relying only on the Stochastic Gradient Decent (SGD) optimiser using a fixed learning rate ($\eta$). 
When using a learning rate of 1e-2, which is a common value used when training a model on a dataset, we can notice that the retrained model incurs catastrophic forgetting and does not improve on the target domain beyond the Source model, even when retraining samples are increased.
Lowering the learning rate overcomes the issue of forgetting and allows the model to adapt gradually to the target domain eventually as the number of samples increases.
Using DIRA improves the issue of forgetting further for low number of samples (i.e. 1 to 2 samples) and allows the model to reach higher accuracies when retraining using less than 10 samples. 
Eventually, the performance of retraining using SGD (with low $\eta$) converges with DIRA as the number of retraining samples increase. % the mode using significantly less number of samples, as little as 1 sample.

Tables \ref{tab:DIRA_CIFAR-10C_results} and \ref{tab:DIRA_CIFAR-100C_results} shows the improvement DIRA achieves upon retraining on 100 samples for each type of corruption in CIFAR-10C and CIFAR-100C benchmarking datasets, respectively, compared to the Source accuracy.

\subsection{Dynamic adaptation scenario for DIRA}
In real-life scenarios, varying domains may occur during operation. 
To visualise how DIRA tackles such a scenario, we plot in Figure~\ref{fig:DynamicScenario} the shift to four different domains consecutively from CIFAR-10C. 
The results depicted show that as soon as samples from the target domain are presented an abrupt improvement occurs in the accuracy of the model. This accuracy continues to grow as more samples from the target domain become present. 

\subsection{Comparison with SOTA}
To assess how well our approach performs compared with SOTA domain adaptation frameworks we compare results with three domain adaptation frameworks from the literature: TTT, NORM and DUA, on ImageNet-C benchmarking dataset. %on our three benchmark datasets. 
Table \ref{tab:DIRA_SOTA_ImageNet_results} show top-1 classification accuracy for the highest severity level on dataset ImageNet-C.
Our DIRA framework performs competitively with SOTA domain adaptation approaches. 
As can be seen from the table, we achieve SOTA overall performance averaged between the different corruptions the dataset. 
This is while using a limited number of samples from the target domain (100 samples).

\section{Conclusions and Future Works} \label{sec:conclusions}
We have introduced a novel domain incremental learning framework, named DIRA (Dynamic Incremental Regularised Adaptation). 
DIRA allows for dynamic adaptation to changing operation environments using a limited number of samples. 
%
% Significant performance gains can be noticed using as little as 5 to 10 samples. 
%
Our approach achieves this using the notion of transfer learning. 
Whereby relevant knowledge from the original domain is retained using regularisation techniques to allow the model to adapt to the target domain making use of transfer learning.
Our DIRA approach proves to be competitive to available domain adaptation approaches in the literature, and achieves SOTA results compared to these approaches.  

% Future works
DIRA is currently categorised as a supervised retraining approach, as it relies on ground truth labels to be provided with samples from the target domain for adaptation. 
This is acceptable but may limit its applications where a source to provide ground truth labels is unavailable. 
Our future work is to explore the combination of DIRA with self-supervised approaches to remove the need for ground truth labels during adaptation.

\printbibliography

@article{Ven2022,
author = {van de Ven, Gido M. and Tuytelaars, Tinne and Tolias, Andreas S},
doi = {10.1038/s42256-022-00568-3},
issn = {25225839},
journal = {Nature Machine Intelligence},
number = {12},
pages = {1185--1197},
publisher = {Springer US},
title = {{Three types of incremental learning}},
volume = {4},
year = {2022}
}

@article{Tao2020,
archivePrefix = {arXiv},
arxivId = {2004.10956},
author = {Tao, Xiaoyu and Hong, Xiaopeng and Chang, Xinyuan and Dong, Songlin and Wei, Xing and Gong, Yihong},
doi = {10.1109/CVPR42600.2020.01220},
eprint = {2004.10956},
issn = {10636919},
journal = {Proceedings of the IEEE Computer Society Conference on Computer Vision and Pattern Recognition},
pages = {12180--12189},
title = {{Few-Shot Class-Incremental Learning}},
year = {2020}
}

@article{JehanzebMirza2022,
archivePrefix = {arXiv},
arxivId = {2204.08817},
author = {{Jehanzeb Mirza}, M. and Masana, Marc and Possegger, Horst and Bischof, Horst},
doi = {10.1109/CVPRW56347.2022.00339},
eprint = {2204.08817},
isbn = {9781665487399},
issn = {21607516},
journal = {IEEE Computer Society Conference on Computer Vision and Pattern Recognition Workshops},
pages = {3000--3010},
title = {{An Efficient Domain-Incremental Learning Approach to Drive in All Weather Conditions}},
volume = {2022-June},
year = {2022}
}

@inproceedings{mirza2022norm,
  title={The norm must go on: dynamic unsupervised domain adaptation by normalization},
  author={Mirza, M Jehanzeb and Micorek, Jakub and Possegger, Horst and Bischof, Horst},
  booktitle={Proceedings of the IEEE/CVF Conference on Computer Vision and Pattern Recognition},
  pages={14765--14775},
  year={2022}
}

@article{Ly2017,
archivePrefix = {arXiv},
arxivId = {1705.01064},
author = {Ly, Alexander and Marsman, Maarten and Verhagen, Josine and Grasman, Raoul P.P.P. and Wagenmakers, Eric Jan},
doi = {10.1016/j.jmp.2017.05.006},
eprint = {1705.01064},
issn = {10960880},
journal = {Journal of Mathematical Psychology},
pages = {40--55},
title = {{A Tutorial on Fisher information}},
volume = {80},
year = {2017}
}

@article{Martens2020,
archivePrefix = {arXiv},
arxivId = {1412.1193},
author = {Martens, James},
eprint = {1412.1193},
issn = {15337928},
journal = {Journal of Machine Learning Research},
pages = {1--76},
title = {{New insights and perspectives on the natural gradient method}},
volume = {21},
year = {2020}
}

@book{Kay1993,
author = {Kay, Steven M.},
pages = {180},
title = {{Fundamentals of Statistical Signal Processing: Estimation Theory}},
year = {1993}
}

@article{Huszar2018,
archivePrefix = {arXiv},
arxivId = {arXiv:1712.03847v1},
author = {Husz{\'{a}}r, Ferenc},
doi = {10.1073/pnas.1717042115},
eprint = {arXiv:1712.03847v1},
issn = {10916490},
journal = {Proceedings of the National Academy of Sciences of the United States of America},
number = {11},
pages = {E2496--E2497},
pmid = {29463735},
title = {{Note on the quadratic penalties in elastic weight consolidation}},
volume = {115},
year = {2018}
}

@Book{GoodBengCour2016,
Title                    = {Deep Learning},
Author                   = {Ian J. Goodfellow and Yoshua Bengio and Aaron Courville},
Publisher                = {MIT Press},
Year                     = {2016},
Address                  = {Cambridge, MA, USA},
Note                     = {\url{http://www.deeplearningbook.org}}
}

@book{MacKay2003,
author = {MacKay, David J. C.},
title = {Information Theory, Inference \& Learning Algorithms},
year = {2002},
isbn = {0521642981},
publisher = {Cambridge University Press},
address = {USA}
}

@article{Li2018c,
archivePrefix = {arXiv},
arxivId = {1606.09282},
author = {Li, Zhizhong and Hoiem, Derek},
doi = {10.1109/TPAMI.2017.2773081},
eprint = {1606.09282},
issn = {19393539},
journal = {IEEE Transactions on Pattern Analysis and Machine Intelligence},
number = {12},
pages = {2935--2947},
pmid = {29990101},
publisher = {IEEE},
title = {{Learning without Forgetting}},
volume = {40},
year = {2018}
}

@article{Zenke2017,
archivePrefix = {arXiv},
arxivId = {1703.04200},
author = {Zenke, Friedemann and Poole, Ben and Ganguli, Surya},
eprint = {1703.04200},
isbn = {9781510855144},
issn = {2640-3498},
journal = {34th International Conference on Machine Learning, ICML 2017},
pages = {6072--6082},
pmid = {31909397},
title = {{Continual learning through synaptic intelligence}},
volume = {8},
year = {2017}
}

@article{harper2021safety,
  title={Safety Validation of Autonomous Vehicles using Assertion-based Oracles},
  author={Harper, Christopher and Chance, Greg and Ghobrial, Abanoub and Alam, Saquib and Pipe, Tony and Eder, Kerstin},
  journal={arXiv preprint arXiv:2111.04611},
  year={2021}
}

@article{eder2021complete,
  title={Complete Agent-driven Model-based System Testing for Autonomous Systems},
  author={Eder, Kerstin I and Huang, Wen-ling and Peleska, Jan},
  journal={arXiv preprint arXiv:2110.12586},
  year={2021}
}

@inproceedings{chance2020agency,
  title={An agency-directed approach to test generation for simulation-based autonomous vehicle verification},
  author={Chance, Greg and Ghobrial, Abanoub and Lemaignan, Severin and Pipe, Tony and Eder, Kerstin},
  booktitle={2020 IEEE International Conference On Artificial Intelligence Testing (AITest)},
  pages={31--38},
  year={2020},
  organization={IEEE}
}

@book{RR-1478-RC,
author="Kalra, Nidhi and Susan M. Paddock",
title="Driving to Safety: How Many Miles of Driving Would It Take to Demonstrate Autonomous Vehicle Reliability?",
address="Santa Monica, CA",
year="2016",
doi="10.7249/RR1478",
publisher="RAND Corporation"
}

@article{Zhang2020,
archivePrefix = {arXiv},
arxivId = {1906.10742},
author = {Zhang, Jie M and Harman, Mark and Ma, Lei and Liu, Yang},
doi = {10.1109/tse.2019.2962027},
eprint = {1906.10742},
issn = {0098-5589},
journal = {IEEE Transactions on Software Engineering},
pages = {1--1},
title = {{Machine Learning Testing: Survey, Landscapes and Horizons}},
year = {2020}
}

@article{Barr2015,
author = {Barr, Earl T. and Harman, Mark and McMinn, Phil and Shahbaz, Muzammil and Yoo, Shin},
doi = {10.1109/TSE.2014.2372785},
issn = {00985589},
journal = {IEEE Transactions on Software Engineering},
number = {5},
pages = {507--525},
publisher = {IEEE},
title = {{The oracle problem in software testing: A survey}},
volume = {41},
year = {2015}
}

@article{Goodfellow2014,
archivePrefix = {arXiv},
arxivId = {1312.6211},
author = {Goodfellow, Ian J. and Mirza, Mehdi and Xiao, Da and Courville, Aaron and Bengio, Yoshua},
eprint = {1312.6211},
journal = {2nd International Conference on Learning Representations, ICLR 2014 - Conference Track Proceedings},
title = {{An empirical investigation of catastrophic forgetting in gradient-based neural networks}},
year = {2014}
}

@article{Kemker2018a,
archivePrefix = {arXiv},
arxivId = {1708.02072},
author = {Kemker, Ronald and McClure, Marc and Abitino, Angelina and Hayes, Tyler L. and Kanan, Christopher},
eprint = {1708.02072},
isbn = {9781577358008},
journal = {32nd AAAI Conference on Artificial Intelligence, AAAI 2018},
pages = {3390--3398},
title = {{Measuring catastrophic forgetting in neural networks}},
year = {2018}
}

@article{Hond2020,
author = {Hond, Darryl and Asgari, Hamid and Jeffery, Daniel},
doi = {10.1109/ICMLA51294.2020.00027},
isbn = {9781728184708},
journal = {Proceedings - 19th IEEE International Conference on Machine Learning and Applications, ICMLA 2020},
pages = {115--121},
title = {{Verifying Artificial Neural Network Classifier Performance Using Dataset Dissimilarity Measures}},
year = {2020}
}

@article{Schaffer2017,
author = {Schaffer, J. David and Land, Walker H.},
doi = {10.1016/j.procs.2017.09.061},
issn = {18770509},
journal = {Procedia Computer Science},
pages = {200--207},
publisher = {Elsevier B.V.},
title = {{Predicting with Confidence: Classifiers that Know What They Don't Know}},
url = {https://doi.org/10.1016/j.procs.2017.09.061},
volume = {114},
year = {2017}
}

@article{Mandelbaum2017,
archivePrefix = {arXiv},
arxivId = {1709.09844},
author = {Mandelbaum, Amit and Weinshall, Daphna},
eprint = {1709.09844},
title = {{Distance-based Confidence Score for Neural Network Classifiers}},
url = {http://arxiv.org/abs/1709.09844},
year = {2017}
}

@article{Xing2019,
archivePrefix = {arXiv},
arxivId = {1912.01730},
author = {Xing, Chen and Arik, Sercan and Zhang, Zizhao and Pfister, Tomas},
eprint = {1912.01730},
pages = {1--12},
title = {{Distance-Based Learning from Errors for Confidence Calibration}},
url = {http://arxiv.org/abs/1912.01730},
year = {2019}
}

@article{deng2012mnist, 
  title={The mnist database of handwritten digit images for machine learning research}, 
  author={Deng, Li}, 
  journal={IEEE Signal Processing Magazine}, 
  volume={29}, 
  number={6}, 
  pages={141--142}, 
  year={2012}, 
  publisher={IEEE} 
}

@article{Xiao2017,
archivePrefix = {arXiv},
arxivId = {1708.07747},
author = {Xiao, Han and Rasul, Kashif and Vollgraf, Roland},
eprint = {1708.07747},
pages = {1--6},
title = {{Fashion-MNIST: a Novel Image Dataset for Benchmarking Machine Learning Algorithms}},
url = {http://arxiv.org/abs/1708.07747},
year = {2017}
}

@book{Koopman2020,
author = {Koopman, Philip and Wagner, Michael},
doi = {10.1007/978-3-030-55583-2_26},
isbn = {9783030555832},
pages = {351--357},
publisher = {Springer International Publishing},
title = {{Positive Trust Balance for Self-driving Car Deployment}},
url = {http://dx.doi.org/10.1007/978-3-030-55583-2_26},
volume = {2},
year = {2020}
}

@misc{chance2023assessing,
      title={Assessing Trustworthiness of Autonomous Systems}, 
      author={Gregory Chance and Dhaminda B. Abeywickrama and Beckett LeClair and Owen Kerr and Kerstin Eder},
      year={2023},
      eprint={2305.03411},
      archivePrefix={arXiv},
      primaryClass={cs.AI}
}

@article{hendrycks2019benchmarking,
  title={Benchmarking neural network robustness to common corruptions and perturbations},
  author={Hendrycks, Dan and Dietterich, Thomas},
  journal={ICLR},
  year={2019}
}

@inproceedings{gepperth2016incremental,
  title={Incremental learning algorithms and applications},
  author={Gepperth, Alexander and Hammer, Barbara},
  booktitle={European symposium on artificial neural networks (ESANN)},
  year={2016}
}

@article{lesort2021understanding,
  title={Understanding continual learning settings with data distribution drift analysis},
  author={Lesort, Timoth{\'e}e and Caccia, Massimo and Rish, Irina},
  journal={arXiv preprint arXiv:2104.01678},
  year={2021}
}

@article{zeno2018task,
  title={Task agnostic continual learning using online variational bayes},
  author={Zeno, Chen and Golan, Itay and Hoffer, Elad and Soudry, Daniel},
  journal={arXiv preprint arXiv:1803.10123},
  year={2018}
}

@inproceedings{li2022energy,
  title={Energy-based models for continual learning},
  author={Li, Shuang and Du, Yilun and van de Ven, Gido and Mordatch, Igor},
  booktitle={Conference on Lifelong Learning Agents},
  pages={1--22},
  year={2022},
  organization={PMLR}
}

@inproceedings{rebuffi2017icarl,
  title={icarl: Incremental classifier and representation learning},
  author={Rebuffi, Sylvestre-Alvise and Kolesnikov, Alexander and Sperl, Georg and Lampert, Christoph H},
  booktitle={Proceedings of the IEEE conference on Computer Vision and Pattern Recognition},
  pages={2001--2010},
  year={2017}
}

@inproceedings{ruvolo2013ella,
  title={ELLA: An efficient lifelong learning algorithm},
  author={Ruvolo, Paul and Eaton, Eric},
  booktitle={International conference on machine learning},
  pages={507--515},
  year={2013},
  organization={PMLR}
}

@article{masse2018alleviating,
  title={Alleviating catastrophic forgetting using context-dependent gating and synaptic stabilization},
  author={Masse, Nicolas Y and Grant, Gregory D and Freedman, David J},
  journal={Proceedings of the National Academy of Sciences},
  volume={115},
  number={44},
  pages={E10467--E10475},
  year={2018},
  publisher={National Acad Sciences}
}

@article{ramesh2021model,
  title={Model Zoo: A Growing" Brain" That Learns Continually},
  author={Ramesh, Rahul and Chaudhari, Pratik},
  journal={arXiv preprint arXiv:2106.03027},
  year={2021}
}

@article{ke2021classic,
  title={CLASSIC: Continual and contrastive learning of aspect sentiment classification tasks},
  author={Ke, Zixuan and Liu, Bing and Xu, Hu and Shu, Lei},
  journal={arXiv preprint arXiv:2112.02714},
  year={2021}
}

@inproceedings{sun2020test,
  title={Test-time training with self-supervision for generalization under distribution shifts},
  author={Sun, Yu and Wang, Xiaolong and Liu, Zhuang and Miller, John and Efros, Alexei and Hardt, Moritz},
  booktitle={International conference on machine learning},
  pages={9229--9248},
  year={2020},
  organization={PMLR}
}

@article{nado2020evaluating,
  title={Evaluating prediction-time batch normalization for robustness under covariate shift},
  author={Nado, Zachary and Padhy, Shreyas and Sculley, D and D'Amour, Alexander and Lakshminarayanan, Balaji and Snoek, Jasper},
  journal={arXiv preprint arXiv:2006.10963},
  year={2020}
}

@article{schneider2020improving,
  title={Improving robustness against common corruptions by covariate shift adaptation},
  author={Schneider, Steffen and Rusak, Evgenia and Eck, Luisa and Bringmann, Oliver and Brendel, Wieland and Bethge, Matthias},
  journal={Advances in neural information processing systems},
  volume={33},
  pages={11539--11551},
  year={2020}
}

@article{kirkpatrick2017overcoming,
  title={Overcoming catastrophic forgetting in neural networks},
  author={Kirkpatrick, James and Pascanu, Razvan and Rabinowitz, Neil and Veness, Joel and Desjardins, Guillaume and Rusu, Andrei A and Milan, Kieran and Quan, John and Ramalho, Tiago and Grabska-Barwinska, Agnieszka and others},
  journal={Proceedings of the national academy of sciences},
  volume={114},
  number={13},
  pages={3521--3526},
  year={2017},
  publisher={National Acad Sciences}
}

@inproceedings{deng2009imagenet,
  title={Imagenet: A large-scale hierarchical image database},
  author={Deng, Jia and Dong, Wei and Socher, Richard and Li, Li-Jia and Li, Kai and Fei-Fei, Li},
  booktitle={2009 IEEE conference on computer vision and pattern recognition},
  pages={248--255},
  year={2009},
  organization={Ieee}
}

@article{krizhevsky2009learning,
  title={Learning multiple layers of features from tiny images},
  author={Krizhevsky, Alex and Hinton, Geoffrey and others},
  year={2009},
  publisher={Toronto, ON, Canada}
}

@article{sun2019unsupervised,
  title={Unsupervised domain adaptation through self-supervision},
  author={Sun, Yu and Tzeng, Eric and Darrell, Trevor and Efros, Alexei A},
  journal={arXiv preprint arXiv:1909.11825},
  year={2019}
}

@inproceedings{maria2017autodial,
  title={Autodial: Automatic domain alignment layers},
  author={Maria Carlucci, Fabio and Porzi, Lorenzo and Caputo, Barbara and Ricci, Elisa and Rota Bulo, Samuel},
  booktitle={Proceedings of the IEEE international conference on computer vision},
  pages={5067--5075},
  year={2017}
}

@inproceedings{he2016deep,
  title={Deep residual learning for image recognition},
  author={He, Kaiming and Zhang, Xiangyu and Ren, Shaoqing and Sun, Jian},
  booktitle={Proceedings of the IEEE conference on computer vision and pattern recognition},
  pages={770--778},
  year={2016}
}

@INPROCEEDINGS{ghobrial2023trustworthiness,
  author={Ghobrial, Abanoub and Hond, Darryl and Asgari, Hamid and Eder, Kerstin},
  booktitle={2023 IEEE International Conference On Artificial Intelligence Testing (AITest)}, 
  title={A Trustworthiness Score to Evaluate DNN Predictions}, 
  year={2023},
  volume={},
  number={},
  pages={9-16},
  doi={10.1109/AITest58265.2023.00011}}

@article{paszke2019pytorch,
  title={Pytorch: An imperative style, high-performance deep learning library},
  author={Paszke, Adam and Gross, Sam and Massa, Francisco and Lerer, Adam and Bradbury, James and Chanan, Gregory and Killeen, Trevor and Lin, Zeming and Gimelshein, Natalia and Antiga, Luca and others},
  journal={Advances in neural information processing systems},
  volume={32},
  year={2019}
}

@article{ghobrial2023evaluation,
  title={Evaluation Metrics for DNNs Compression},
  author={Ghobrial, Abanoub and Balemans, Dieter and Asgari, Hamid and Reiter, Phil and Eder, Kerstin},
  journal={arXiv preprint arXiv:2305.10616},
  year={2023}
}

\end{document}